\documentclass[sn-mathphys,Numbered]{sn-jnl}

\usepackage{graphicx}%
\usepackage{amsmath,amssymb,amsfonts}%
\usepackage{amsthm}%
\usepackage{mathrsfs}%
\usepackage[title]{appendix}%
\usepackage{xcolor}%
\usepackage{textcomp}%
\usepackage{manyfoot}%
\usepackage{multirow}
\usepackage{booktabs}%
\usepackage{algorithm}%
\usepackage{algorithmicx}%
\usepackage{algpseudocode}%
\usepackage{listings}%
\usepackage{hyperref}%

\theoremstyle{thmstyleone}%
\theoremstyle{thmstyletwo}%

\theoremstyle{thmstylethree}%

\begin{document}

\title[Article Title]{1D-Convolutional transformer for Parkinson disease diagnosis from gait}


\author*[1]{\fnm{Safwen} \sur{Naimi}}\email{safwen.naimi@teluq.ca}

\author[1]{\fnm{Wassim} \sur{Bouachir}}\email{wassim.bouachir@teluq.ca}

\author[2]{\fnm{Guillaume-Alexandre} \sur{Bilodeau}}\email{gabilodeau@polymtl.ca}

\affil[1]{\orgdiv{Data Science Laboratory}, \orgname{University of Quebec (TELUQ)}, \city{Montréal}, \country{Canada}}

\affil[2]{\orgdiv{LITIV lab.}, \orgname{Polytechnique Montréal}, \city{Montréal}, \country{Canada}}


\abstract{This paper presents an efficient deep neural network model for diagnosing Parkinson's disease from gait. More specifically, we introduce a hybrid ConvNet-Transformer architecture to accurately diagnose the disease by detecting the severity stage. The proposed architecture exploits the strengths of both Convolutional Neural Networks and Transformers in a single end-to-end model, where the former is able to extract relevant local features from Vertical Ground Reaction Force (VGRF) signal, while the latter allows to capture long-term spatio-temporal dependencies in data. In this manner, our hybrid architecture achieves an improved performance compared to using either models individually. Our experimental results show that our approach is effective for detecting the different stages of Parkinson's disease from gait data, with a final accuracy of 88\%, outperforming other state-of-the-art AI methods on the Physionet gait dataset. Moreover, our method can be generalized and adapted for other classification problems to jointly address the feature relevance and spatio-temporal dependency problems in 1D signals. Our source code and pre-trained models are publicly available at \href{https://github.com/SafwenNaimi/1D-Convolutional-transformer-for-Parkinson-disease-diagnosis-from-gait}{https://github.com/SafwenNaimi/1D-Convolutional-transformer-for-Parkinson-disease-diagnosis-from-gait}.}




\maketitle

\section{Introduction}\label{sec1}

Parkinson's disease is a progressive neurological disorder mainly affecting the ability of the patient to control movement, in addition to the risk of causing mental and cognitive disorders depending on patients and severity stages. It is named after the British doctor James Parkinson, who first described the condition in 1817 \cite{Parkinson1969ANEO}. Parkinson described the characteristic symptoms of the disease, including tremors, stiffness, and difficulty with movement. The exact cause of Parkinson's disease is not known, but it is believed to be related to a combination of genetic and environmental factors. The disease is associated with a loss of cells in a specific area of the brain called the \textit{substantia nigra}, which is involved in the production of a neurotransmitter called dopamine. The loss of dopamine in the brain leads to the characteristic symptoms of Parkinson's disease. Over the years, many treatments have been developed for Parkinson's disease, including medications, surgery, and physical therapy \cite{Stoker2020RecentDI}. Despite these treatments, there is no cure for the condition, and it is generally a chronic and progressive disease. In recent years, research into the causes and potential treatments for Parkinson's disease has intensified, and new discoveries are being made at a regular pace. As a result, the outlook for people with Parkinson's disease has improved, and many are able to live full and active lives with the condition.

Staging Parkinson's disease relates to the process of determining its severity in a patient. It is a way to classify the progression of the disease and help guide treatment decisions. There are several ways to stage Parkinson's disease, including clinical evaluation surveys, such as the Unified Parkinson's Disease Rating Scale (UPDRS) \cite{Goetz2008MovementDS} and the Hoehn and Yahr (H\&Y) scale \cite{Bhidayasiri2012ParkinsonsDH}. Both methods are based on surveys and are carried out by a healthcare professional. The UPDRS is a commonly used scale for evaluating the severity of PD symptoms. The evaluation is done through a series of questions and tasks that are designed to assess various aspects of the disease, including motor symptoms, non-motor symptoms, and overall functional ability. The UPDRS is an important tool in the management of Parkinson's disease, providing crucial information to both patients and healthcare professionals \cite{2003TheUP}. The Hoehn and Yahr (H\&Y) scale is another tool that is often used to assess the severity of PD. This scale is based on a clinical examination and is used to determine the stage of the disease. The scale ranges from stage 1 (mild symptoms) to stage 5 (severe symptoms), and the results are used to guide treatment decisions and to help predict the course of the disease \cite{Hoehn1998ParkinsonismOP}.

Several decision-support methods have been proposed to detect Parkinson's disease. These methods aim to assist in the early detection and diagnosis of PD by identifying characteristic patterns or abnormalities in signals that are indicative of the disease. Generally, the proposed methods learn to classify input data and find patterns that indicate PD. Some of them involve designing and implementing algorithms to extract handcrafted features from a person's gait or other movement data that are indicative of the disease. Spectral analysis \cite{Heida2013PowerSD, Patel2021SpectralAO} and wavelet analysis \cite{Chang2022EvaluatingTD, Rizvi2019EarlyDO} are examples of handcrafted signal processing methods that have been used to detect Parkinson's disease by decomposing signals into frequency components. They can be useful in identifying characteristic changes in the frequency content of physiological signals, such as Electroencephalogram (EEG) and Electromyography (EMG) that are indicative of the disease. Time-domain analysis is another approach that involves analyzing the time-domain representation of a signal, such as a person's gait data, to identify specific features that may be indicative of PD. Changes in stride length or walking speed over time can be used as features for detecting the disease \cite{Ertugrul2016DetectionOP, Xia2015ClassificationOG, Mannini2016AML}.

Handcrafted features can be used as inputs to train machine learning algorithms such as support vector machines (SVMs) \cite{Burges1998ATO}, decision trees, and deep learning algorithms to classify a given signal as belonging to a PD patient or a healthy individual. However, handcrafted methods are often limited in their ability to extract distinctive representations. To address this, other methods use deep learning architectures to learn relevant features directly from the signal as they are able to better represent the distinctive characteristics of the disease \cite{Nanni2017HandcraftedVN}.

In this paper, we focus on how to diagnose PD while predicting disease severity. Previous works \cite{Caramia2018IMUBasedCO, Veeraragavan2020ParkinsonsDD, Machi2019Deep1F} have shown that it is possible to analyze foot signals, captured during a walk, to detect characteristic patterns or abnormalities related to PD. In several studies, the signals collected from the feet of a patient represent the vertical ground reaction force (VGRF) in function of time, as measured by 18-foot sensors \cite{Keller1996RelationshipBV, Takahashi2004VerticalGR, Muniz2006PrincipalCA}. The VGRF is the force exerted by the ground on the feet of a person as they walk, and it can provide valuable information about the patient's gait and physical condition. Therefore, in this work, we are using 1-dimensional (1D) VGRF signals to classify the stage of PD based on the H\&Y scale. 

Our method relies on a novel hybrid end-to-end architecture, where a Convolutional Neural Network (ConvNet) is used to extract relevant local features from VGRF signals, followed by Transformers to align these features, in order to classify the stage of PD. To ensure a precise diagnosis, the signals are first divided into segments. These segments are then classified by the model into corresponding PD stages. The final patient classification is determined by a majority voting process, which takes into account the most occurring stage among all segments of the patient's walk.

The key contribution of this work is a new neural network architecture designed for detecting PD and staging disease severity. Our Hybrid ConvNet-Transformer comprises two main components, to capture both local and global features from signal. The ConvNet captures local patterns, while the Transformer captures long-term dependencies and temporal relationships. By exploiting these two aspects, our model can learn complex relationships in the signal that are indicative of PD. Through experiments, we demonstrate that this method is more accurate than existing methods for detecting and staging PD. Moreover, the proposed architecture can be generalized for other signal classification problems, to jointly address the feature relevance and spatio-temparal dependency issues. The source code of this project is publicly available to ensure reproducibility for future research.

\section{Related Work}\label{sec2}

Previous research on PD stage classification has used various machine learning and deep learning models to analyze gait data, which can be collected using wearable sensors or vision-based systems. These models have been used to investigate the potential of gait data for accurately classifying PD patients into different stages of the disease. Caramia et al. \cite{Caramia2018IMUBasedCO} studied the use of inertial sensors, consisting of 8 Inertial Measurement Units (IMUs), to collect gait data from 25 patients with Parkinson's disease (PD). They extracted spatio-temporal gait features from the data, which were then used as input for different classifiers. Mirelman et al. \cite{Mirelman2021DetectingSM} also used IMU to classify PD stages based on H\&Y scale. They applied a RUSBoost classifier to the data. However, the authors noted that the classification rates of PD stages with IMU were not satisfactory, due to issues such as sensor drift and noise contamination. This means that the accuracy of these classifiers can be affected by changes in the sensor over time and by noise in the data, leading to lower classification rates. 

The use of computer vision techniques to predict the severity of PD from markerless RGB / RGBD cameras has gained attention. These techniques involve extracting 2D/3D poses of the PD patient from videos and then using machine learning or deep learning models to classify the PD stages based on the extracted poses. Sabo et al. \cite{Sabo2020AssessmentOP} used this approach to extract both 2D and 3D skeletons of PD patients from videos and applied multivariate ordinal Logistic Regression (LR) models to classify the patients based on their poses. While this approach showed some promise, it has been found that the results are not always accurate. Further research is thus needed to improve the detection of Parkinson's disease from videos and more advanced techniques are needed to achieve better results.

El Maachi et al. \cite{Machi2019Deep1F} and Veeraragavan et al. \cite{Veeraragavan2020ParkinsonsDD} studied the use of machine learning techniques to classify Parkinson's disease stages from gait data. El Maachi et al. used a 1D-Convolutional Neural Network (1D-ConvNet) to stage PD, while Veeraragavan et al. used an Artificial Neural Network (ANN). Both methods achieved good performance in staging PD based on the physiological data. However, there are still some limitations in using only ConvNets to capture the relationship between sensors. ConvNets are known to be effective in capturing local spatial information, but they are not the best option for capturing global patterns or relationships between different sensors. On the other hand, using only ANNs to capture local information may not be sufficient, as it may not be able to fully capture the complex patterns in physiological data.
Additionally, both methods may not be able to effectively handle missing or noisy data, and may not be able to fully capture the complex temporal patterns in physiological data. These limitations highlight the need for further research and development in this field, such as utilizing more advanced architectures or incorporating domain knowledge. Our work is in line with this research axis, with the objective of improving the robustness of PD diagnosis and the accuracy of severity prediction.
\begin{figure}[h]
	\centering
	\includegraphics[scale=0.378]{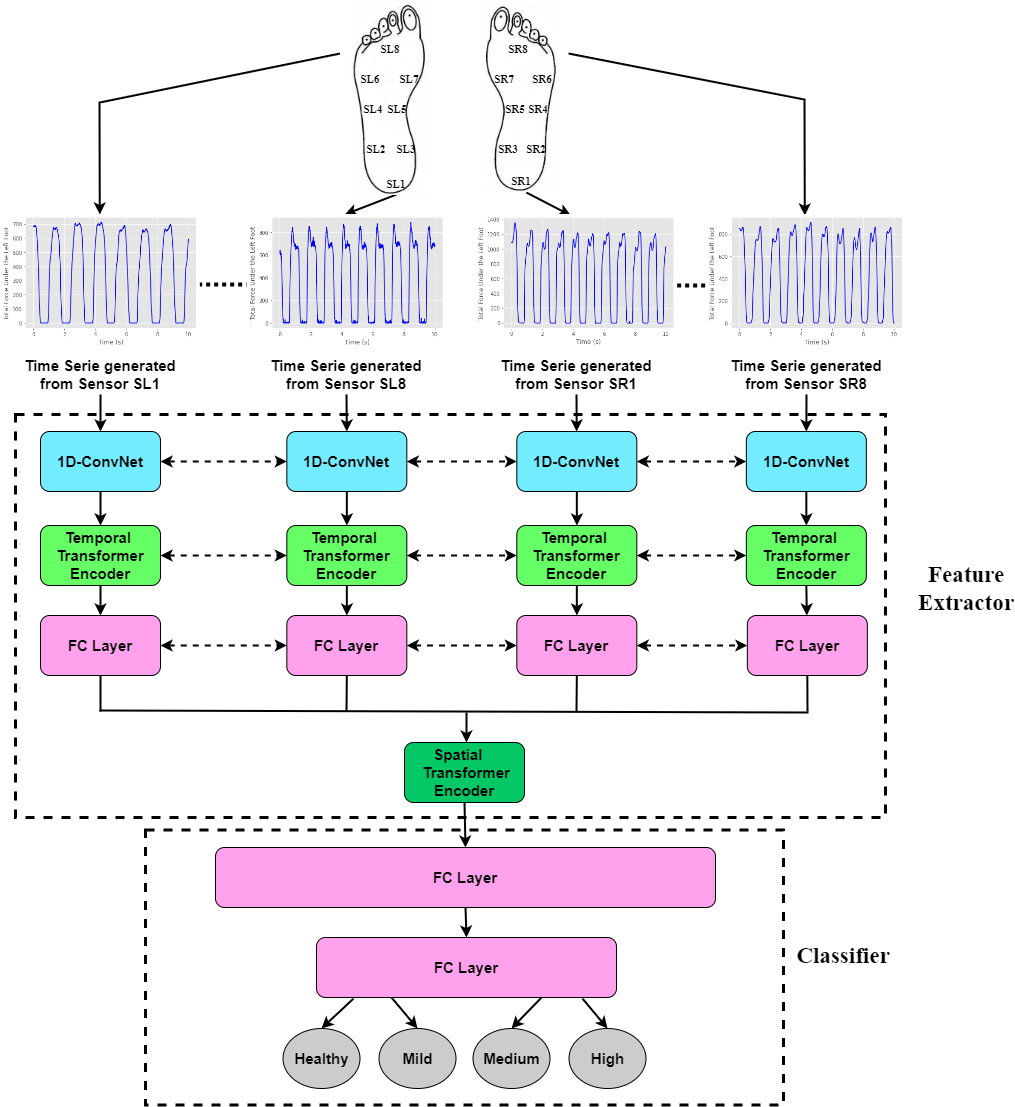}
	\caption{Overall Architecture of our Hybrid ConvNet-Transformer model. It is made of a Feature Extractor (1D-ConvNet/Temporal Transformer Encoder/FC Layer/Spatial Transformer Encoder) and a Classifier (Two FC Layers followed by an Output Layer divided into the four stages of PD).}
	\label{FIG:1}
\end{figure}

\section{Proposed Approach}\label{sec3}
In this study, we address the issue of identifying Parkinson's disease stages using gait information from foot sensors. The overall architecture of our proposed model is depicted in Figure \ref{FIG:1}. Our input data are $S$ VGRF 1D signals from a patient's walk, which represent the vertical ground reaction force as a function of time recorded by foot sensors. We first divided each walk into smaller segments of $p$ elements to obtain more data. The number of elements is chosen so that enough information is stored in each segment.

The feature extractor is made up of $S$ parallel 1D-ConvNets. Each network accepts a segment as input and processes it through convolutional layers. This 1D-ConvNet parallelization enables the independent treatment of every signal, which is recommended since they have different characteristics because each sensor collects a specific VGRF from a specific position of the foot. Therefore, each time series has its own deep features. 

The temporal dependencies, which are the connection between two values of a vector spaced apart in time, can be captured using the Transformers. The output vector of each 1D-ConvNet is thus passed through a temporal transformer encoder. The spatial dependencies between each set of vectors emanating from the foot sensors can likewise be captured using the spatial transformers after performing a dimension reduction. Such dependencies can be the relative position between sensors and co-occurring VGRFs. The classifier is a fully connected network operating on the concatenation of the features extracted by the spatial transformer encoder. The final walk classification is decided according to the majority classification of all the subject segments. Our proposed architecture is detailed in the following subsections.

\subsection{1D-ConvNet}\label{subsec2}
The network starts with $S$ parallel 1D-ConvNet layers, designed to analyze the gait patterns of each foot sensor by identifying patterns over time in the time series data, thereby allowing the extraction of distinct features. This is because each sensor records data from a specific point, resulting in different time series with diverse deep features that need to be analyzed separately. 

\begin{figure}[b]%
        \centering
		\includegraphics[scale=0.45]{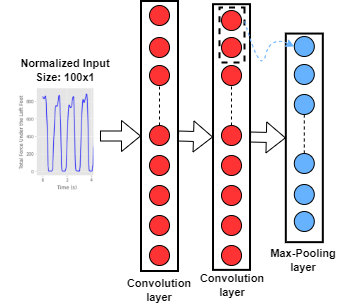}
	\caption{Internal structure of a 1D-ConvNet block}
	\label{FIG:5}
\end{figure}

The model is made up of $B_c$ convolutional blocks, each consisting of 2 convolutional layers followed by 1 max pooling layer, as illustrated in Figure \ref{FIG:5}.
This design is used to extract important features and reduce the computational complexity \cite{Park2020LowcomplexityCW}. The 1D-convolutional layers apply filters to the 1D-gait VGRF signals, creating a feature map that summarizes the detected features. The convolution is applied using the following equation:
\begin{equation}
y[i] = \sum_{j=0}^{n-1} x[i+j]*w[j] + b,
\end{equation}
where $x$ is the input data, $w$ is the kernel (or filter), $b$ is the bias, and $y$ is the output. The symbol $*$ represents the element-wise multiplication, and the sum is taken over the filter window of size $n$. 

The advantage of this approach is that it takes into account the local structure of the data while maintaining the sequential information of the input sequence and learning local correlations simultaneously. By adjusting the number of convolutional blocks and the size of the filters, it is possible to adjust the level of abstraction and the robustness of the model.

\subsection{Temporal Transformer Encoder}
In our model, we employ temporal transformer encoder architecture to reduce intraclass variance and capture long-term dependencies in the data. The encoder is a crucial component of the temporal transformer model. It is responsible for encoding the input sequence and generating a set of representations that capture the dependencies within the data \cite{Lohit2019TemporalTN}. We opted for $S$ parallel temporal transformer encoders, each containing $B_{{tt}}$ blocks to enhance the performance of the model. Each temporal transformer encoder block in our architecture is composed of one multi-head attention layer with two heads and a feed-forward network, similar to the architecture proposed in BERT for natural language processing \cite{Tyagi2014AdvancesIN}.

To capture temporal dependencies, we used a fixed positional encoding with a constant step according to the appropriate segment length. This is motivated by the fact that gait data was separated into fixed and constant segments. Moreover, the use of a fixed positional encoder allowed for effective modeling of the temporal dependencies within these segments. The fixed positional encoder used in our model is a variation of the original transformer positional encoder. It is designed to work with fixed-length inputs. Additionally, we normalized the positional encoding to prevent the complete masking of information in the original vector.

\subsection{Spatial Transformer Encoder}
The main role of the spatial transformer is to identify dependencies between sensors, by taking into account the spatial distribution of sensors on the foot and potentially discovering correlations between them. 

The outputs of the $S$ parallel temporal transformer encoders are concatenated after performing dimension reduction with the $S$ parallel Fully Connected Layers. This helps to obtain a more compact data representation and remove any redundant information. This concatenated vector is used as input for the $B_{{st}}$ spatial transformer encoder blocks, which are supplemented with another fixed positional encoder used to provide the spatial transformer encoder with information about the relative position of the input elements in the concatenated vector. Similarly to the temporal transformer encoders, we opted for one multi-head attention layer with two heads and a feed-forward network.

\subsection{Classifier Block}
To predict the PD stage, we used a classifier composed of two fully connected layers and an output layer as the final components of our hybrid ConvNet-Transformer model. This block takes in the output of the spatial transformer encoder and produces a probability distribution over classes of PD severity. We trained the classifier using a categorical cross-entropy loss function \cite{LeCun1999ObjectRW} to update the weights and biases. The probabilities were then used to determine the predicted PD stage for each input data.

\section{Experimental results and evaluation}\label{sec4}
In the following subsections, we present the results of our experiments and provide a detailed analysis of our findings. First, we describe the dataset used in our experiments. Next, we explain the evaluation metrics used to measure the performance of our model. We also provide the technical details of our model implementation. We then present the results of our experiments, discuss them in details and compare them to other methods. Finally, we conduct an ablation study to further analyze the impact of different components of our model on performance.
\subsection{Dataset}
\begin{table*}[b]
\tiny
  \centering
  \caption{Demographics of healthy subjects and PD subjects in three datasets}
\begin{tabular}{|p{0.5cm}|p{1cm}|p{0.6cm}|c|c|cc|p{0.7cm}|c|}
\hline
\multicolumn{ 1}{|c|}{{\bf Dataset}} & \multicolumn{ 1}{|c|}{{\bf Group}} & \multicolumn{ 1}{|c|}{{\bf Subjects}} & \multicolumn{ 1}{|c|}{{\bf Male}} & \multicolumn{ 1}{|c|}{{\bf Female}} & \multicolumn{ 2}{|c}{{\bf Age (Yrs)}} & \multicolumn{ 1}{|c|}{{\bf Height
(Meter)}} & \multicolumn{ 1}{|c|}{{\bf Weight
(Kg)}} \\
\multicolumn{ 1}{|c|}{{\bf }} & \multicolumn{ 1}{|c|}{{\bf }} & \multicolumn{ 1}{|c|}{{\bf }} & \multicolumn{ 1}{|c|}{{\bf }} & \multicolumn{ 1}{|c|}{{\bf }} & {\bf Mean±SD} & {\bf Range} & \multicolumn{ 1}{|c|}{{\bf }} & \multicolumn{ 1}{|c|}{{\bf }} \\
\hline
\multicolumn{ 1}{|c|}{\bf Yogev et al. \cite{Yogev2005DualTG}} &    Healthy &         18 &         10 &          8 &   57.9±6.7 &      37-70 &   1.68±.08 &  74.2±12.7 \\
\multicolumn{ 1}{|c|}{} & PD  &         29 &         20 &          9 &   61.6±8.8 &      36-77 &   1.67±.07 &  73.1±11.2 \\
\hline
\multicolumn{ 1}{|c|}{\bf Hausdorff et al. \cite{Hausdorff2007RhythmicAS}} &    Healthy &         26 &         12 &         14 & 39.31±18.51 &      20-74 &   1.83±.08 & 66.8±11.07 \\

\multicolumn{ 1}{|c|}{} & PD  &         29 &         16 &         13 & 66.80±10.85 &      44-80 &   1.87±.15 & 75.1±16.89 \\
\hline
\multicolumn{ 1}{|c|}{\bf Toledo et al. \cite{FrenkelToledo2005TreadmillWA}} &    Healthy &         29 &         18 &         11 &   64.5±6.8 &      53-77 &   1.69±.08 &  71.5±11.0 \\

\multicolumn{ 1}{|c|}{} & PD  &         35 &         22 &         13 &   67.2±9.1 &      61-84 &   1.66±.07 &   70.3±8.4 \\
\hline
\end{tabular}  
  \label{tab:tab1}%
\end{table*}%
For our study, we used the publically available Physionet gait dataset \cite{Physionet}.
The gait dataset was created by three groups of researchers, namely 
Yogev et al. \cite{Yogev2005DualTG}, Hausdorff et al. \cite{ Hausdorff2007RhythmicAS} and Toledo et al. \cite{FrenkelToledo2005TreadmillWA}. It contains three gait patterns acquired through walking on a level ground, walking with rhythmic auditory stimulation (RAS), and walking on a treadmill. The dataset contains the gait pattern from 93 patients affected with PD and 73 healthy subjects. The dataset collected by \cite{Yogev2005DualTG} consists of a gait pattern for normal walking on a level ground. The dataset contributed by \cite{Hausdorff2007RhythmicAS} contains the gait cycle for walking at a comfortable pace with RAS. The contribution from \cite{FrenkelToledo2005TreadmillWA} comprises a gait time series data for walking on a treadmill.

Table \ref{tab:tab1} gives the demographics of the participating subjects, from whom gait data was collected. Table \ref{tab:tab2} presents the total number of healthy and PD subjects with their level of severity in each dataset, determined according to the H\&Y scale.
\begin{table*}[t]
 \centering
  \caption{Number of subjects in the three datasets based on severity rating}
\small
\begin{tabular}{|r|r|r|r|r|}
\hline
   \bf Dataset &    \bf Healthy & \bf Severity 2 & \bf Severity 2.5 & \bf Severity 3 \\
\hline
       \bf Yogev et al. \cite{Yogev2005DualTG}  &     18 &    15 &     8 &   6 \\
\hline
       \bf Hausdorff et al. \cite{ Hausdorff2007RhythmicAS}  &     26 &    12 &    13 &   4 \\
\hline
       \bf Toledo et al. \cite{FrenkelToledo2005TreadmillWA}  &     29 &    29 &     6 &   0 \\
\hline
\end{tabular}  
   \label{tab:tab2}%
\end{table*}%

\begin{figure*}
     \centering
		\includegraphics[scale=0.6]{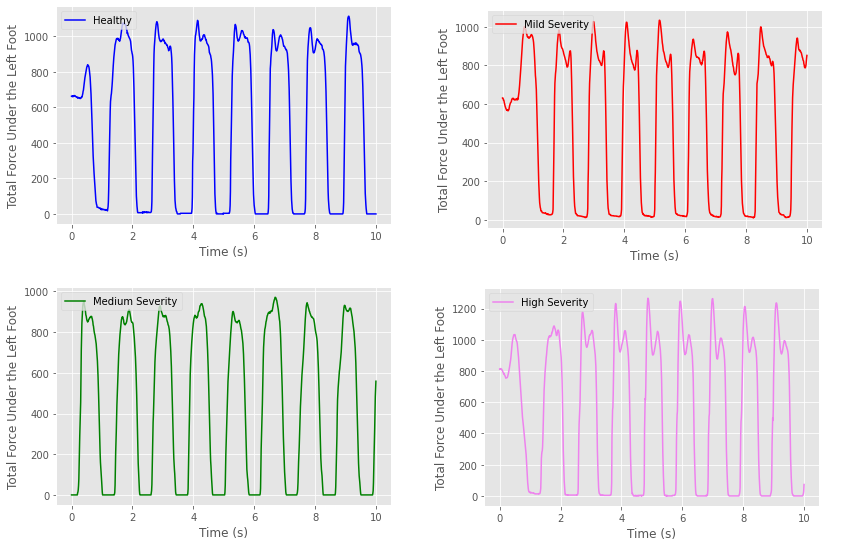}
	\caption{Plots of the left foot VGRF signal of a Control patient and patients with different Parkinson’s disease stages}
	\label{FIG:100}
\end{figure*}

Figure \ref{FIG:100} illustrates the difference in VGRF reading for control versus PD subjects. The plots of the right foot VGRF signal of a control patient and a Parkinson's disease patient may differ in terms of amplitude, shape, and signal timing. The recurrence plots of Control subjects display orderly and consistent texture patterns. The VGRF signal in these plots has a steady amplitude and a regular shape, with predictable timing. As the severity of Parkinson's disease increases, the deviations in the VGRF signal may become more pronounced. The VGRF signal shows greater asymmetry in the gait pattern and reduced force applied to the left foot compared to a control patient. These deviations may impact mobility and functional ability to a greater extent and may require more intensive treatment and management.

\subsection{Evaluation Metrics}
We tested our algorithm on 300 walks using 10-Fold cross-validation. At the subject level, we separated both the Parkinson's and the control groups into 10 folds. As a result, we were able to maintain the same dataset balance for each fold (70\% Parkinson - 30\% Control).
The utilized performance metrics are given below, and include $TP$ as the number of true positives, $TN$ as the number of true negatives, $FP$ as the number of false positives, and $FN$ as the number of false negatives.

\begin{equation}
{  \textbf{Accuracy (\%):} } Acc =\frac{TP+TN}{TP+TN+FP+FN}\times 100\%
\end{equation}
\begin{equation}
 {  \textbf{Precision:} } Pr=\frac{TP}{TP+FP}
\end{equation}
\begin{equation}
 {  \textbf{Recall:} } Re=\frac{TP}{TP+FN}
\end{equation}
\begin{equation}
 {  \textbf{F1-Score} } =2\times \frac{ Pr \times Re}{ Pr+ Re}
\end{equation}

The accuracy shows the degree of all truly classified observations. Precision is a measure of how many of the positive predictions made are correct (true positives). The recall is a measure of how many of the positive cases the classifier correctly predicted, over all the positive cases in the data. The F1-score may be thought of as a harmonic mean of accuracy and recall, with the highest value being 1 and the lowest being 0.

\subsection{Implementation details}
Before analyzing the 1D signals extracted from the Physionet gait dataset, we applied data preprocessing techniques. For the control patients, we noticed that some data values were missing and replaced them with zeroes. We also normalized the signals to have a mean of zero and a standard deviation of one, and divided each walk into smaller segments of $p=100$ time steps with 50\% overlap. These segments were labeled with the subject category and were initially created from the 1D signals.

The 1D-ConvNet layers are made of $B_c=2$ blocks. Since we are processing data from 18 sensors, we use $S=18$ parallel processing streams. The parameters of each 1D-ConvNet layer are given in Table \ref{tab:tab5}.
\begin{table}[t]
  \centering
  \caption{1D-ConvNet parameters used in our approach}
\begin{tabular}{|c|c|c|c|c|}
\hline
{\bf Layer} & {\bf Type} & {\bf Stride} & {\bf
Kernels} & {\bf Padding} \\
\hline
   {\bf 1} & Convolutive &          1 &          8 &    'valid' \\
\hline
   {\bf 2} & Convolutive &          1 &         16 &    'valid' \\
\hline
   {\bf 3} & Max-Pooling &          2 &         16 &    'valid' \\
\hline
   {\bf 4} & Convolutive &          1 &         16 &    'valid' \\
\hline
   {\bf 5} & Convolutive &          1 &         16 &    'valid' \\
\hline
   {\bf 6} & Max-Pooling &          2 &         16 &    'valid' \\
\hline
\end{tabular}  
  \label{tab:tab5}%
\end{table}%

The temporal transformer encoder is made of $B_{{tt}}=1$ block. Its input is a vector of $p=100$ elements resulting from the 1D-ConvNet layers, and the output is a vector of 10 elements obtained by applying global average pooling, a dropout regularization layer, and a FC layer. This last FC layer is used to scale down the output vectors from 100 elements to 10 elements to reduce its dimensionality and make it easier to extract meaningful information from data. The spatial transformer encoder is made of $B_{{st}}=1$ block. The corresponding input is composed of 18 outputs of 10 elements from the $S=18$ parallel temporal transformers that have been dimensionally reduced. Except for the output, where we utilized a Softmax activation function, every fully connected layer is using the SeLU activation function (scaled exponential linear units) \cite{Pratama2020TrainableAF}.

The proposed hybrid ConvNet-Transformer model is trained, validated, and tested separately, using a batch size of 150 samples for each iteration. We trained for 30 epochs.  The proposed model is trained using the Nadam \cite{Dozat2016IncorporatingNM} stochastic optimization method with the following parameters: $\alpha$ = 0.001, $\beta_1$ = 0.9, $\beta_2$ = 0.999, where $\alpha$ is the learning rate, $\beta_1$ and $\beta_2$ are the exponential decay rates for the first and second-moment estimations, respectively.
To improve the model performance and reduce overfitting, we opted for a dropout rate of 0.1 and early stopping.

\subsection{Results and Discussion}
In Table \ref{tab:tab11}, we compare the results of our proposed architecture with those of other studies to stage PD. Based on the results of this table, our hybrid ConvNet-Transformer achieved the highest accuracy among the compared methods. It was able to effectively classify different stages of Parkinson's disease, resulting in an accuracy of 87.89\%. This suggests that the combination of the ConvNet and Transformer can successfully detect and classify Parkinson's disease by taking advantage of their strengths.

The support vector machine with a radial basis function kernel (SVM-RBF) algorithm used in the study of Caramia et al. \cite{Caramia2018IMUBasedCO} had the lowest accuracy of 75.6\% when evaluated using the H\&Y scale. This method is a type of supervised learning algorithm that can be used for classification tasks. It works by finding a hyperplane in feature space that maximally separates different classes. SVM-RBFs are often effective for classification tasks but can be sensitive to the choice of kernel parameters. The study conducted by El Maachi et al. \cite{Machi2019Deep1F} achieved the second-highest accuracy among the four selected studies, using a 1D-ConvNet detection algorithm only and UPDRS scale stages. In this study, the ConvNet method was able to identify and classify different stages of Parkinson's disease, resulting in an accuracy of 85.30\%, although not as accurate as our hybrid ConvNet-Transformer architecture. Finally, the study by Veeraragavan et al. \cite{Veeraragavan2020ParkinsonsDD} had an accuracy of 76.08\% for 10-fold cross-validation, using an ANN detection algorithm and H\&Y scale. This could be due to the limitations of the ANN algorithm in accurately detecting and classifying the stages of Parkinson's disease. Overall, these results indicate that the use of our hybrid ConvNet-Transformer model improved the accuracy of detecting PD and classifying its severity stages.

\begin{figure}
        \centering
		\includegraphics[scale=0.6]{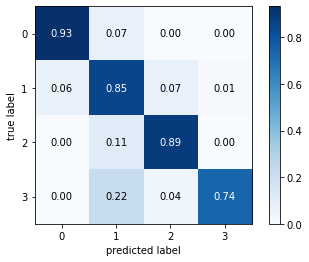}
	\caption{Confusion Matrix for the H\&Y staging. 0: Healthy (Severity 0), 1: Mild (Severity 2), 2: Medium (Severity 2.5), 3: High (Severity 3)}
	\label{FIG:8}
\end{figure}

\begin{table*}[t]
  \centering
  \caption{Comparison of Classification Algorithms for Parkinson's Disease}
\begin{tabular}{|c|c|c|c|}
\hline
{\bf Selected Study} & {\bf Detection Algorithm} & {\bf Stages} & {\bf Accuracy} \\
\hline
Caramia et al. \cite{Caramia2018IMUBasedCO} &     SVM-RBF &      H\&Y Scale &    75.60\% \\
\hline
Veeraragavan et al. \cite{Veeraragavan2020ParkinsonsDD}  &        ANN &  H\&Y Scale &    76.08\% \\
\hline
El Maachi et al. \cite{Machi2019Deep1F} &     1D-ConvNet &      UPDRS &    85.30\% \\
\hline
Our Work & Hybrid ConvNet-Transformer  &  H\&Y Scale &    \bf 87.89\% \\
\hline
\end{tabular}  
  \label{tab:tab11}%
\end{table*}%
The detailed confusion matrix obtained from the 10-Fold Cross-Validation method of our architecture is displayed in Figure \ref{FIG:8}. It has an accuracy of 93\% for class 0 (Healthy), an accuracy of 85\% for class 1 (H\&Y stage 2), an accuracy of 89\% for class 2 (H\&Y stage 2.5), and an accuracy of 74\% for class 3 (H\&Y stage 3). Overall, the classifier seems to perform well for the majority classes (class 0, class 1, and class 2), but less for the minority class (class 3). This might be attributable to the dataset imbalance, which can unfairly distort results because more data are available for stages 2 and 2.5 than for stage 3 in the Physionet gait dataset as mentioned in Table \ref{tab:tab2}.

The global accuracy rate of our classifier is 87.89\%, which corresponds to the classification of a patient based on his/her entire walk in the validation set. It is worth noting that some healthy individuals may walk in a way that appears similar to the characteristic shuffling gait often seen in people with Parkinson's disease. This can lead to incorrect classifications in our algorithm, as it may mistake a healthy person's abnormal walk for a Parkinsonian walk. Detailed 10-Fold Cross-Validation results of the severity level prediction per class are shown in Table \ref{tab:tab6} below. 
\begin{table}[t]
  \caption{10-Fold Cross-Validation results for the H\&Y staging}
  \centering
\small
\begin{tabular}{|c|c|c|c|c|c|}
\hline
{\bf Label} & {\bf \#Subjects} & {\bf Precision} & {\bf Recall} & {\bf F1-Score} &  {\bf Accuracy}    \\
\hline
{\bf 0 \textit{(Healthy)}} & 90 &       0.80 &       0.93 &       0.86 &  \multirow{4}{*}{\centering 0.88}  \\
{\bf 1 \textit{(Severity 2)}} &      110 &       0.86 &       0.85 &       0.85 &   \\
{\bf 2 \textit{(Severity 2.5)}}&      73 &       0.88 &       0.89 &       0.88 &    \\
{\bf 3 \textit{(Severity 3)}}&     27 &       0.95 &       0.74 &       0.83 &     \\
\hline
\hline
{\bf Macro Avg}&      300 &       0.87 &       0.85 &       0.86 &   \\
{\bf Weighted Avg}&      300 &       0.87 &       0.88 &       0.88 &   \\
\hline
\end{tabular}  
\label{tab:tab6}%
\end{table}

Our model has higher precision and recall for the severity 2 (label 1) and severity 2.5 (label 2) classes. For the severity 3 (label 3) class, the model has a slightly lower recall but a relatively high precision. For the Healthy (label 0) class, the model has a high recall but a relatively low precision. Overall, the macro average and weighted average F1-score of the model are both around 0.87, indicating that the model performs well in general. We can observe that the F1-score is quite stable across classes, indicating that the algorithm predictions are generally accurate. 

\begin{figure}[t]
        \centering
		\includegraphics[scale=0.75]{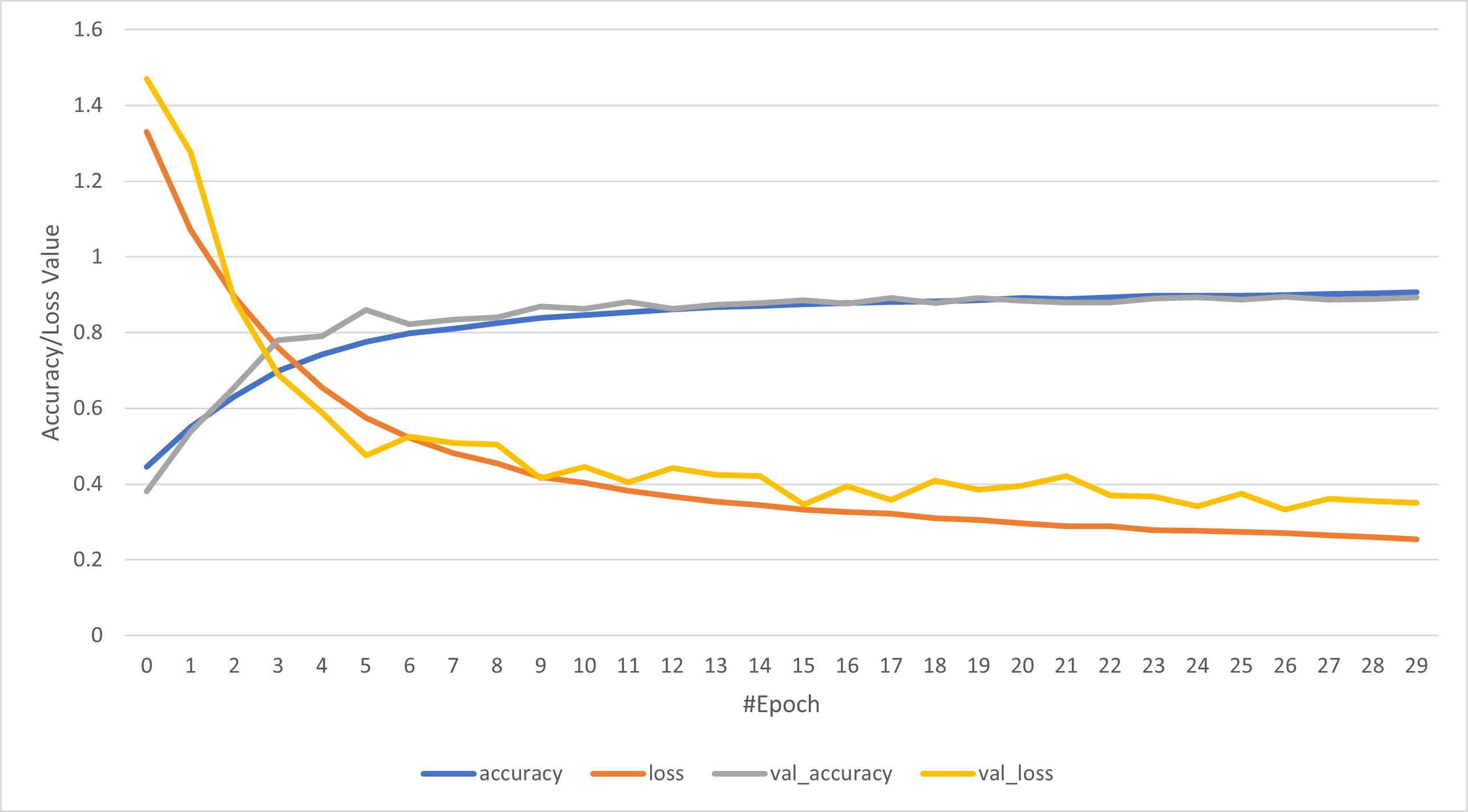}
	\caption{Training and Validation Plots for Parkinson’s severity level prediction}
	\label{FIG:9}
\end{figure}

Figure \ref{FIG:9} illustrates the accuracy and loss for the training and validation plots. We can observe that we were able to prevent overfitting between training and validation. As the training accuracy improves, the accuracy of the validation set also improves. Additionally, the same pattern is observed for the loss curves, as the training loss decreases, the validation loss also decreases, indicating a strong correlation between the two.

\subsection{Ablation Study}
In this section, we investigate ablation scenarios by comparing the proposed hybrid ConvNet-Transformer model to some closely related model designs in order to identify its important properties. By conducting an ablation study, we can gain a better understanding of how different components of the architecture contribute to the overall performance of the model and identify potential improvements that could be made to enhance its accuracy.

Table \ref{tab:tab35} shows the principal components of each model used in the ablation study. In model A, we removed the Temporal Transformer Blocks and kept the rest of the components. In model B, we removed the Spatial Transformer Block and the previous parallel fully connected layers. In model C, we removed the Transformer part (Temporal Transformer Encoder and Spatial Transformer Encoder) and kept the 1D-ConvNet Block. The obtained architecture is similar to that in \cite{Machi2019Deep1F}. In model D, we removed the 1D-ConvNet and we maintained all the transformer blocks. We obtained a similar model to that in \cite{Nguyen2022TransformersF1}.
\begin{table*}[t]
  \caption{Summary of the ablation study. Boldface is used to highlight the best results}
\centering
\tiny
\begin{tabular}{|p{1.7cm}|p{1.5cm}|p{1.5cm}|p{1.6cm}|c|c|c|c|c|}
\hline
           & \bf 1D-ConvNet & \bf Spatial Transformer & \bf Temporal Transformer &      \bf Label &  \bf Precision &     \bf Recall &   \bf F1-Score &   \bf Accuracy  \\
\hline
\multirow{5}{*}{\textbf{Model A}}  & \multicolumn{ 1}{|c|}{\multirow{5}{*}{\checkmark}} & \multicolumn{ 1}{|c|}{\multirow{5}{*}{\checkmark}} & \multicolumn{ 1}{|c|}{\multirow{5}{*}{\texttimes}} &    \emph{Healthy} & \bf 0.89 &        0.8 &       0.84 & \multirow{5}{*}{\centering 0.83} \\

\multicolumn{ 1}{|c}{} & \multicolumn{ 1}{|c|}{} & \multicolumn{ 1}{|c|}{} & \multicolumn{ 1}{|c|}{} &  \emph {Severity 2} &       0.77 & {\bf 0.92} &       0.84 &  \\

\multicolumn{ 1}{|c}{} & \multicolumn{ 1}{|c|}{} & \multicolumn{ 1}{|c|}{} & \multicolumn{ 1}{|c|}{} & \emph {Severity 2.5} &       \bf0.89 &       0.81 &       0.85 & \\

\multicolumn{ 1}{|c}{} & \multicolumn{ 1}{|c|}{} & \multicolumn{ 1}{|c|}{} & \multicolumn{ 1}{|c|}{} & \emph{Severity 3} &       0.91 &       0.52 &       0.68 & \\

\multicolumn{ 1}{|c}{} & \multicolumn{ 1}{|c|}{} & \multicolumn{ 1}{|c|}{} & \multicolumn{ 1}{|c|}{} &  {Weighted Avg} &       0.85 &       0.83 &       0.83 & \\
\hline
\multirow{5}{*}{\textbf{Model B}} & \multicolumn{ 1}{|c|}{\multirow{5}{*}{\checkmark}} & \multicolumn{ 1}{|c|}{\multirow{5}{*}{\texttimes}} & \multicolumn{ 1}{|c|}{\multirow{5}{*}{\checkmark}} &    \emph{Healthy} &       0.72 &       0.93 &       0.81 & \multirow{5}{*}{\centering 0.84}  \\

\multicolumn{ 1}{|c}{} & \multicolumn{ 1}{|c|}{} & \multicolumn{ 1}{|c|}{} & \multicolumn{ 1}{|c|}{} & \emph{Severity 2} &       0.85 &       0.84 &       \bf0.85 &  \\

\multicolumn{ 1}{|c}{} & \multicolumn{ 1}{|c|}{} & \multicolumn{ 1}{|c|}{} & \multicolumn{ 1}{|c|}{} & \emph{Severity 2.5} &       0.85 &       0.87 &       0.86 & \\

\multicolumn{ 1}{|c}{} & \multicolumn{ 1}{|c|}{} & \multicolumn{ 1}{|c|}{} & \multicolumn{ 1}{|c|}{} & \emph{Severity 3} &       0.94 &        0.6 &       0.73 & \\

\multicolumn{ 1}{|c}{} & \multicolumn{ 1}{|c|}{} & \multicolumn{ 1}{|c|}{} & \multicolumn{ 1}{|c|}{} & {Weighted Avg} &       0.85 &       0.84 &       0.84 & \\
\hline
\multirow{5}{*}{\textbf{Model C}} & \multicolumn{ 1}{|c|}{\multirow{5}{*}{\checkmark}} & \multicolumn{ 1}{|c|}{\multirow{5}{*}{\texttimes}} & \multicolumn{ 1}{|c|}{\multirow{5}{*}{\texttimes}} &    \emph{Healthy} &       0.75 &        0.9 &       0.82 & \multirow{5}{*}{\centering 0.84} \\

\multicolumn{ 1}{|c}{} & \multicolumn{ 1}{|c|}{} & \multicolumn{ 1}{|c|}{} & \multicolumn{ 1}{|c|}{} & \emph{Severity 2} &       0.84 &       0.86 &       \bf0.85 & \\

\multicolumn{ 1}{|c}{} & \multicolumn{ 1}{|c|}{} & \multicolumn{ 1}{|c|}{} & \multicolumn{ 1}{|c|}{} & \emph{Severity 2.5} &       0.87 &       0.86 &       0.87 & \\

\multicolumn{ 1}{|c}{} & \multicolumn{ 1}{|c|}{} & \multicolumn{ 1}{|c|}{} & \multicolumn{ 1}{|c|}{} & \emph{Severity 3} &        0.9 &       0.67 &       0.77 & \\

\multicolumn{ 1}{|c}{} & \multicolumn{ 1}{|c|}{} & \multicolumn{ 1}{|c|}{} & \multicolumn{ 1}{|c|}{} & Weighted Avg &       0.85 &       0.84 &       0.84 & \\
\hline
\multirow{5}{*}{\textbf{Model D}} & \multicolumn{ 1}{|c|}{\multirow{5}{*}{\texttimes}} & \multicolumn{ 1}{|c|}{\multirow{5}{*}{\checkmark}} & \multicolumn{ 1}{|c|}{\multirow{5}{*}{\checkmark}} &    \emph{Healthy} &       0.58 & {\bf 0.97} &       0.72 & \multirow{5}{*}{\centering 0.79} \\

\multicolumn{ 1}{|c}{} & \multicolumn{ 1}{|c|}{} & \multicolumn{ 1}{|c|}{} & \multicolumn{ 1}{|c|}{} & \emph{Severity 2} &       0.82 &       0.81 &       0.82 & \\

\multicolumn{ 1}{|c}{} & \multicolumn{ 1}{|c|}{} & \multicolumn{ 1}{|c|}{} & \multicolumn{ 1}{|c|}{} & \emph{Severity 2.5} &       0.88 &       0.71 &       0.78 & \\

\multicolumn{ 1}{|c}{} & \multicolumn{ 1}{|c|}{} & \multicolumn{ 1}{|c|}{} & \multicolumn{ 1}{|c|}{} & \emph{Severity 3} &        0.9 &       0.73 &       0.82 & \\

\multicolumn{ 1}{|c}{} & \multicolumn{ 1}{|c|}{} & \multicolumn{ 1}{|c|}{} & \multicolumn{ 1}{|c|}{} & {Weighted Avg} &       0.82 &       0.79 &       0.79 & \\
\hline
\multirow{5}{*}{\textbf{Final Model}} & \multicolumn{ 1}{|c|}{\multirow{5}{*}{\checkmark}} & \multicolumn{ 1}{|c|}{\multirow{5}{*}{\checkmark}} & \multicolumn{ 1}{|c|}{\multirow{5}{*}{\checkmark}} &    \emph{Healthy} &        0.8 &       0.93 & {\bf 0.86} & \multirow{5}{*}{\centering \textbf{0.88}} \\

\multicolumn{ 1}{|c}{} & \multicolumn{ 1}{|c|}{} & \multicolumn{ 1}{|c|}{} & \multicolumn{ 1}{|c|}{} & \emph{Severity 2} & {\bf 0.86} &       0.85 & {\bf 0.85} & \\

\multicolumn{ 1}{|c}{} & \multicolumn{ 1}{|c|}{} & \multicolumn{ 1}{|c|}{} & \multicolumn{ 1}{|c|}{} & \emph{Severity 2.5} & {0.88} & {\bf 0.89} & {\bf 0.88} & \\

\multicolumn{ 1}{|c}{} & \multicolumn{ 1}{|c|}{} & \multicolumn{ 1}{|c|}{} & \multicolumn{ 1}{|c|}{} & \emph{Severity 3} & {\bf 0.95} & {\bf 0.74} & {\bf 0.83} & \\

\multicolumn{ 1}{|c}{} & \multicolumn{ 1}{|c|}{} & \multicolumn{ 1}{|c|}{} & \multicolumn{ 1}{|c|}{} & {Weighted Avg} & {\bf 0.87} & {\bf 0.88} & {\bf 0.88} & \\
\hline
\end{tabular}  
\label{tab:tab35}%
\end{table*}

Based on a comparison between the results of the final model and those of other versions, we can observe that our final model achieved the best overall performance. While the efficiency of different models varies across different labels, our model generally performs better across most labels. This demonstrates that the proposed hybrid ConvNet-Transformer architecture is able to ensure better stability of the model performance across different labels, resulting in a more robust model overall. This ablation study shows that removing or replacing components of our model leads to a decrease in performance. This demonstrates that each component of our model is important to ensure optimal overall performance.

\section{Conclusion}\label{sec5}

In this paper, we proposed a method for detecting Parkinson's disease stages using a novel hybrid ConvNet-Transformer architecture. By using a Convolutional Neural Network to extract local features and a Transformer Network to capture long-range dependencies, our method is able to effectively model the complex nature and dynamics of the parkinsonian gait. We evaluated our method on the public Physionet gait dataset using VGRF data and demonstrated its effectiveness in detecting the different stages of the disease. Compared to existing methods, our approach showed an improved performance with an accuracy of 87.89\%. The increasing availability of biomedical sensors represents an important opportunity for widespread implementation of our approach in the future, particularly for monitoring gait abnormalities in elderly populations. In conclusion, our hybrid ConvNet-Transformer architecture marks an advancement towards the development of AI-based tools for Parkinson's disease diagnosis. The source code and pre-trained models for this study are publicly available on GitHub, making it possible for other researchers to reproduce the results and build upon our work.

\section*{Declarations}

\begin{itemize}
\item Funding: This work was supported by research grants from the Natural Sciences and Engineering Research Council of Canada (NSERC) [Wassim Bouachir: Discovery Grant number RGPIN-2020-04937, Guillaume-Alexandre Bilodeau: Discovery Grant number RGPIN-2020-04633].
\item Conflict of interest/Competing interests: Not applicable.
\item Ethics approval: Not applicable.
\item Consent to participate: Not applicable.
\item Consent for publication: Not applicable.
\item Availability of data and materials: The dataset analysed during the current study is available at \url{https://physionet.org/content/gaitpdb/1.0.0/}.
\item Code availability: The code is publicly available at:
\href{https://github.com/SafwenNaimi/1D-Convolutional-transformer-for-Parkinson-disease-diagnosis-from-gait}{https://github.com/SafwenNaimi/1D-Convolutional-transformer-for-Parkinson-disease-diagnosis-from-gait-patterns}.
\item Authors' contributions: 

\textbf{Safwen Naimi}: Data curation, Formal analysis, Methodology, Software, Validation, Writing – original draft.

\textbf{Wassim Bouachir}:  Conceptualization, Funding acquisition, Methodology, Resources, Supervision, Writing – review \& editing.

\textbf{Guillaume-Alexandre Bilodeau}: Conceptualization, Funding acquisition, Methodology, Supervision, Writing – review \& editing.
\end{itemize}

\bibliography{sn-bibliography}

\end{document}